\documentclass[letterpaper, 10 pt, conference]{ieeeconf}  

\IEEEoverridecommandlockouts                              

\usepackage{amsfonts}
\usepackage{amsmath,amssymb,amsfonts}
\usepackage{url}
\usepackage{etoolbox}
\usepackage{graphicx}
\usepackage{optidef}
\usepackage{algpseudocode}
\usepackage{mathtools,amssymb}
\usepackage{svg}
\usepackage{ulem}
\usepackage{stix}
\usepackage{amsmath,mleftright,mathtools}
\DeclareMathAlphabet\mathbfcal{OMS}{cmsy}{b}{n}
\usepackage{hyperref}
\usepackage{stfloats}
\usepackage{lipsum}
\usepackage{comment}

\usepackage[linesnumbered,ruled,vlined,commentsnumbered]{algorithm2e}

\usepackage{colortbl}

\usepackage{fancyhdr}
\usepackage{tikz}
\usepackage{leftindex}

\DeclareMathOperator*{\argmin}{arg\,min} 

\title{\LARGE \bf
Safety-Aware Optimal Scheduling for Autonomous Masonry Construction using Collaborative Heterogeneous Aerial Robots 
}

\author{Marios-Nektarios Stamatopoulos, Shridhar Velhal, Avijit Banerjee and George Nikolakopoulos
\thanks{The Authors are with the Robotics and Artificial Intelligence Group, Department of Computer, Electrical and Space Engineering, Lule\r{a} University of Technology, 971 87 Lule\r{a}, Sweden}%
\thanks{Cooresrponing author's e-mail: \tt\small{marsta@ltu.se}}
}

\begin{document}
\maketitle
\begin{tikzpicture}[overlay, remember picture]
    \node[anchor=north,
          xshift=0.0cm,
          yshift=-0.15cm, align=center]
         at (current page.north)
         {\fontsize{9}{10}\selectfont This paper has been accepted for publication at the \\2025 IEEE/RSJ International Conference on Intelligent Robots and Systems (IROS 2025).
};
\end{tikzpicture}
\vspace{-8mm}

\thispagestyle{empty}
\pagestyle{empty}

\begin{abstract}
This paper presents a novel high-level task planning and optimal coordination framework for autonomous masonry construction, using a team of heterogeneous aerial robotic workers, consisting of agents with separate skills for brick placement and mortar application.
This introduces new challenges in scheduling and coordination, particularly due to the mortar curing deadline required for structural bonding and ensuring the safety constraints among UAVs operating in parallel.
To address this, an automated pipeline generates the wall construction plan based on the available bricks while identifying static structural dependencies and potential conflicts for safe operation.
The proposed framework optimizes UAV task allocation and execution timing by incorporating dynamically coupled precedence deadline constraints that account for the curing process and static structural dependency constraints, while enforcing spatio-temporal constraints to prevent collisions and ensure safety. 
The primary objective of the scheduler is to minimize the overall construction makespan while minimizing logistics, traveling time between tasks, and the curing time to maintain both adhesion quality and safe workspace separation.
The effectiveness of the proposed method in achieving coordinated and time-efficient aerial masonry construction is extensively validated through Gazebo simulated missions. The results demonstrate the framework's capability to streamline UAV operations, ensuring both structural integrity and safety during the construction process.
- A video with the framework and the simulated mission is available at \href{https://youtu.be/kGvFGDCUkDQ}{https://youtu.be/kGvFGDCUkDQ}

\end{abstract}

\section{Introduction}
Recent advancements in robotic autonomy, driven by sophisticated sensors and planning tools, are paving the way for a shift in the construction landscape \cite{advancementsRoboticsConstr}. Robotics has emerged as a transformative technology, offering the potential to enhance productivity and improve occupational safety \cite{saidi2016}, while addressing autonomous coordination with high safety risks and reducing human workload.
Researchers and practitioners are actively exploring various gantry systems, 
utilizing both additive manufacturing concepts 
\cite{largeScale3DPrint} 
and traditional bricklaying approaches \cite{BRIX, inSituFabr}. Additionally, ground-based autonomous robots and mobile robots equipped with manipulators have demonstrated their potential to enhance construction workflows, whether by operating on elevated moving rails \cite{brickLabyrinth} or extruding cement while in motion \cite{printingWhileMoving}.
Despite their effectiveness and efficiency, such robotic systems have significant limitations including restricted reach in critical and infrastructure deficient areas, due to their lack of flexibility, as well as extensive logistics required for deployment. An intriguing alternative is the use of Unmanned Aerial Vehicles (UAVs) as autonomous construction workers. Following the additive manufacturing approaches, these UAVs can function as airborne robotic builders, extruding material while flying \cite{nature_aerial_AM,stamatopoulosChunkingJINT,stamatopoulosICUAS,Stamatopoulos2024Experiment}.
Alternatively, UAVs can be used for autonomous execution of pick up and place operation for stacking bricks to assemble large-scale architectures \cite{flightAssembledArch}. 
\begin{figure}[t]
    \centering
    \includegraphics[width=\linewidth]{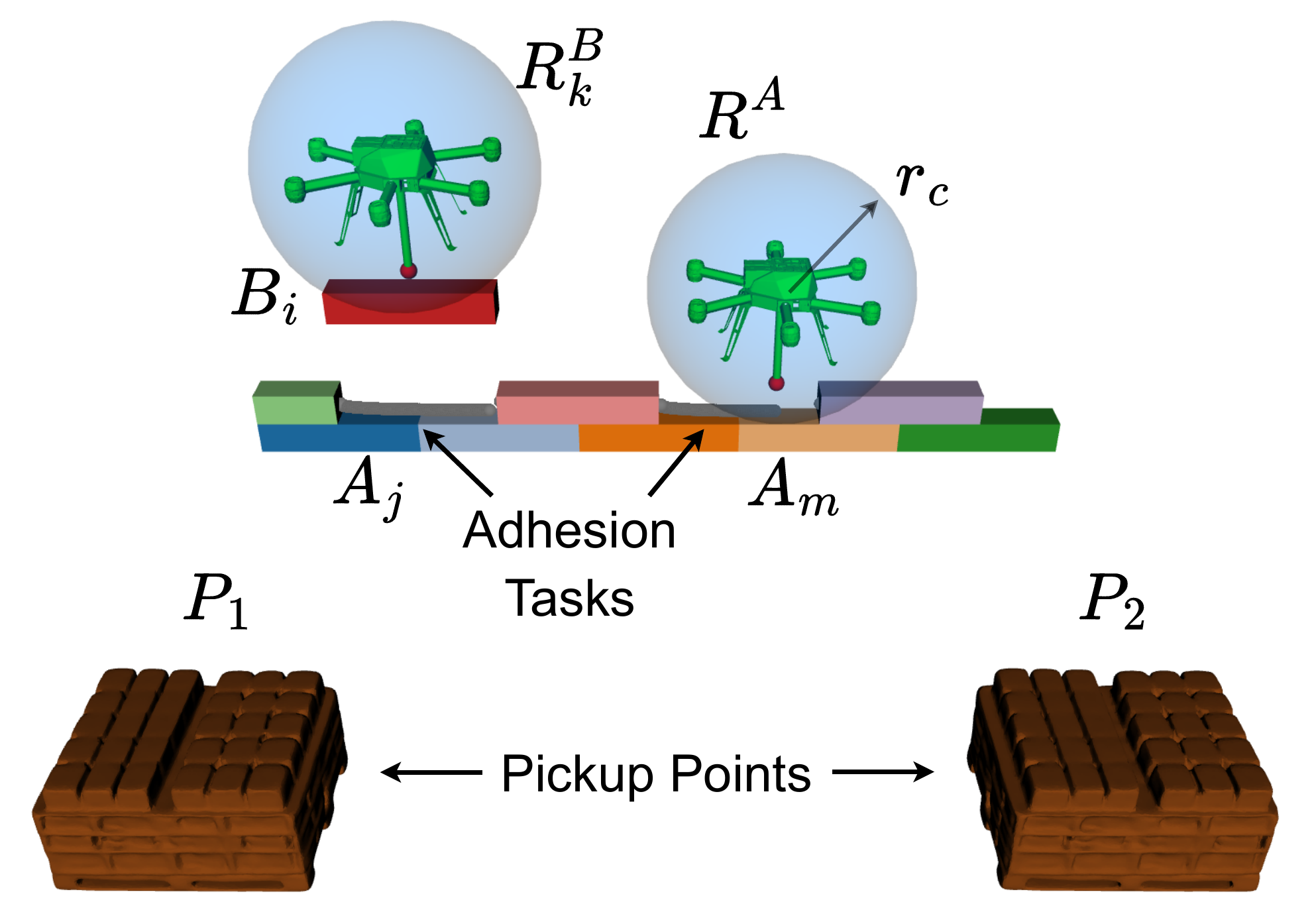}
    \caption{Concept figure of the proposed framework where a brick pick-place robot $R^B_k$ is placing the brick $B_i$ while the robot $R^A$ is executing the adhesion task $A_m$. The minimum clearance $r_c$ is visualized via blue transparent spheres while brick pickup points $P_k$ are shown on the bottom.}
    \label{fig:concept}
    \vspace{-6mm}
\end{figure}
While early stage foundational development in UAV-based autonomous bricklaying has been explored in the literature \cite{feasibility}, the scope of research has remained limited to brick pick-up and placement, primarily focusing on aerial manipulation mechanisms. However, for full-scale autonomous construction, simply stacking bricks is not enough since structural integrity requires the application of cement-like materials to reinforce adhesion and bonding. Towards this, the article introduces a novel heterogeneous multi-agent aerial robotic fleet, where one class of UAVs is responsible for precise brick placement, while another is designated to carry and spray adhesive material such as mortar, while seamlessly integrating traditional bricklaying techniques with additive manufacturing principles.

\subsection{Related Work}
The concept of cooperative wall building using multiple UAVs was initially introduced in \cite{flightAssembledArch} where given a construction blueprint with the exact positioning of each brick, the tasks were reactively assigned to the available UAVs from a central module based on their state and battery level while the possible collisions between them are handled reactively by reserving space for their trajectories. The mortar between the bricks was placed manually by a human right before the brick was placed for pickup. 
Additionally, in the second challenge of the MBZIRC 2020 competition, which required collaborative construction using both ground and aerial robots, other researchers also focused on this issue. In \cite{mbzircWallSaska}, each UAV is given specific segments of the wall to build and corresponding stacks of bricks to draw from. This assignment is done in such a way that there’s a division of the construction tasks, which helps in optimizing their movements and reducing the likelihood of overlap or collision during the building process. Additionally in \cite{mbzircMetaheuristicWallplanner}, near-optimal building plans are calculated by modeling the construction process as a variant of the Team Orienteering Problem, incorporating precedence and concurrency constraints to ensure safe and orderly brick placements withing the UAVs and solved using a meta-heuristic method known as the Greedy Randomized Adaptive Search Procedure (GRASP). In \cite{mbzircBarbara}, a hierarchical task planning and coordination framework that enables effective task allocation and scheduling using various brick types is proposed, demonstrating how a mobile robot cooperates with up to three UAVs using a decentralized approach that optimizes the utilization of each robot's capabilities.
%
However, none of the aforementioned literature considers the problem of placing mortar autonomously, including deadline constraints for the brick on the top of adhesion and the associated challenges of planning for heterogeneous UAVs.

\begin{figure*}[b]
    \vspace{-5mm}

    \includegraphics[width=\linewidth]{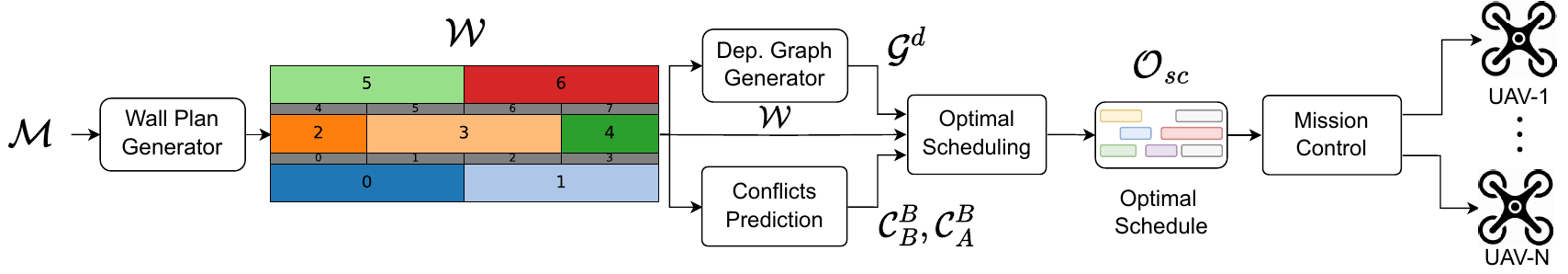}
    \vspace{-5mm}
    \caption{Multi-Agent
Brick and Mortar Autonomous Construction System Architecture Overview.}
    \label{fig:blockDiagram}
    \vspace{-5mm}
\end{figure*}

\subsection{Contributions}
This article presents a novel high level planning framework for optimal scheduling and task allocation in aerial masonry construction using heterogeneous UAVs, where dedicated UAVs autonomously perform brick placement and mortar application as separate tasks. Their coordinated actions are dynamically modeled as part of the optimal planning to enhance construction efficiency while ensuring structural robustness through proper adhesion with a strong focus on minimizing overall makespan (construction time), UAV logistics movements, and accounting for mortar curing duration before brick placement.
A key challenge addressed in this framework is the need for consideration of mortar's curing time, which starts immediately after its application and defines a bounded time window for the subsequent brick placement. To ensure optimal adhesion quality and construction efficiency, this constraint is integrated into the scheduling process as dynamically coupled precedence and deadline constraints.
Additionally, the proposed scheduling framework systematically organizes tasks within a dependency graph to handle static structural constraints while simultaneously enforcing spatio-temporal safety constraints. These include clearance requirements to prevent collisions and ensure safe, parallel operation of UAVs throughout the construction process.
The key contributions of the paper can be listed as:
\begin{itemize} 
    \item Novel framework utilizing heterogeneous aerial workers, wherein a dedicated UAV facilitates the adhesion between bricks through precise mortar application ensuring their safety during simultaneous operation and adhering to structural constraints.
    \item Optimal scheduling that considers the curing time of the adhesive substance, ensuring efficient construction by framing curing time as dynamically coupled precedence-deadline constraint.
    \item Automated construction plan generation pipeline, including the placement of bricks and adhesion tasks,  while systematically identifying static structural dependencies and potential conflicts to enhance overall scheduling efficiency.
\end{itemize}
To the best of the authors' knowledge, this article is the first time the concept of full-scale autonomous aerial construction workers using multiple heterogeneous UAV types—one for brick placement and another for mortar application—has been explored, while addressing the challenges associated with mortar curing deadline.

\section{Problem Statement - Challenges}\label{sec:problStatement}

\begin{figure}
    \centering
    \includegraphics[width=0.92\linewidth]{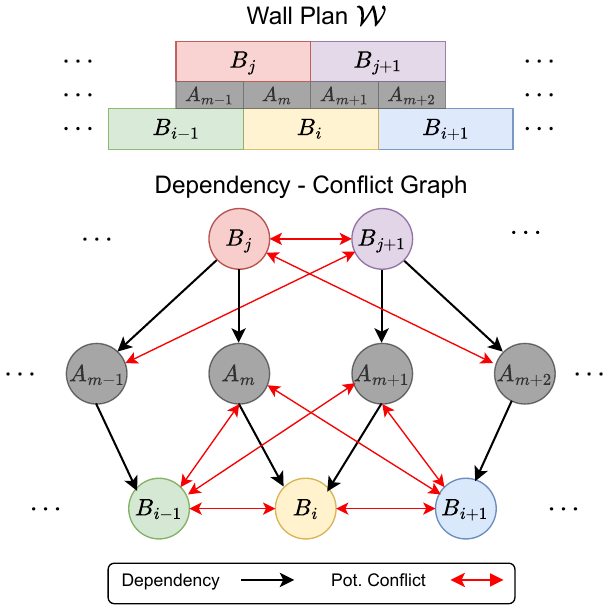}
    \caption{Wall plan blueprint $\mathcal{W}$ depicting the brick tasks $B_i$ (top) and the adhesion tasks $A_m$, along with the corresponding dependency graph $\mathcal{G}^d$ illustrating the structural relationships between tasks (black) and identified potential conflicts (red) $\mathcal{C}^B_B$, $\mathcal{C}^B_A$ between brick and adhesion tasks (bottom).}
    \label{fig:wallPlanDepConflGraphConcept}
    \vspace{-6mm}
\end{figure}

This paper addresses the problem of constructing a wall using multiple heterogeneous UAVs, each assigned one of two distinct roles—brick placement or mortar application—while handling two types of bricks. As shown in Fig. \ref{fig:blockDiagram}, given a 3D mesh $\mathcal{M}$ representing the desired wall and the dimensions of the available bricks, a construction plan $\mathcal{W}$ needs to be generated, specifying the exact placement of each brick $B_i$ and the associated position and length of each adhesion task $A_j$.
The construction process involves a team of UAVs, $\mathcal{R^B} = \{R^B_0, R^B_1, \dots, R^B_N\}$, responsible for picking up and placing bricks, along with a dedicated UAV, $R^A$, that sprays the adhesion material. Each UAV $R^B_i$ operates from a designated pickup location $P_i$ with a sufficient supply of bricks for the entire mission, while the adhesion material, carried by $R^A$, is also assumed to be sufficient.
A major challenge arises due to the dynamically coupled deadline constraints associated with the adhesion task. Once the mortar is applied, the next brick must be placed within a specified curing time, $d_{cur}$, to ensure proper bonding. Failing to meet this time constraint can compromise the structural integrity of the wall, making precise scheduling essential.
In addition to the curing-time constraint, the schedule must also respect structural dependencies, ensuring that bricks and adhesion are placed in a physically valid sequence over time. To ensure a valid construction sequence, these constraints are represented as a dependency graph $\mathcal{G}^d$ as shown in Fig. \ref{fig:wallPlanDepConflGraphConcept} with black connecting arrows.
Apart from these static constraints, the problem becomes more challenging due to the need to maintain a minimum safety clearance $r_c$ between UAVs to ensure safe and conflict-free operation. 
These conflicts might occur both among brick placement tasks and between brick placement and adhesion tasks, captured in sets $\mathcal{C}^B_B$ and $\mathcal{C}^B_A$, respectively (shown by red arrows). This introduces additional complexity to the scheduling process, as task assignments and timing must be adjusted to prevent congestion while maintaining construction efficiency. 
The dependency and safety conditions are fed to the scheduler which assigns UAVs to the tasks with their respective starting times.
The objective is to minimize the overall construction time while ensuring safety and compliance to adhesion constraints while
UAV movements should also be optimized to reduce travel distances and minimize delays between adhesion application and brick placement, thereby maximizing bonding efficiency and construction speed. Finally, the schedule should be sent to mission control to ensure real-time UAV coordination and execution.
\section{Brick-Mortar Multi-task Scheduling Framework}

\subsection{Wall Plan Generation} \label{sec:wallPlanGen}
Towards generating a construction plan $\mathcal{W}$ for the given mesh wall $\mathcal{M}$, its dimensions $w_m, h_m$ and $l_m$ are extracted corresponding to the width, height and length of the wall. In this paper, two different bricks were considered available, the full-bricks with width, height and thickness equal to $w_{fb}, h_{fb}, t_{fb}$ and the half-bricks with the same characteristics but width equal to $w_{hb} = w_{fb}/2$. A predefined running-bond pattern is selected to be followed on the placement of the bricks, which is a common bricklaying technique \cite{runningBond} that relies on alternating bricks in each row, necessitating half bricks at the ends to maintain structural integrity and proper alignment. The algorithm followed to generate the brick layout is shown in Algorithm \ref{alg:brickLayout}.
\vspace{-3mm}
\begin{algorithm}
    \caption{Generate Brick Layout Plan}
    \label{alg:brickLayout}
    \small
    \DontPrintSemicolon
    Initialize grid dims. based on the wall and brick sizes\;
    Create an empty layout matrix $\mathcal{W}$\;
    \For{each row in the grid} {
        \For{each column in the row} {
            \If{row is even and first column} {
                Place a half-brick with an offset\;
            } \Else {
                Place a full brick at the adjusted position\;
            }
            Store brick in the layout $\mathcal{W}$\;
            \If{row is odd and last column} {
                Add a half-brick at the end of the row\;
            }
        }
    }
    \Return layout matrix $\mathcal{W}$\;
\end{algorithm}
\vspace{-6mm}

\subsection{Adhesion - Dependency Graph Calculation}
After the brick layout plan is generated, the dependency graph and adhesion tasks are constructed by analyzing all pairs of bricks to determine structural dependencies. A dependency from brick $B_i$ to brick $B_j$ is defined when $B_i$ is placed on top of $B_j$.
For each dependent pair of bricks, their horizontal overlap is calculated. Based on this overlap, an adhesion task $A_m$ is created at the position of its start, with a horizontal span equal to its width.
Finally, the dependency graph $\mathcal{G}^d$ is updated with two directed edges: $(B_i, A_m)$ and $(A_m, B_j)$, representing the structural relationship between the top and bottom bricks, along with the adhesion task. An overview of the process can be found in Algorithm \ref{alg:depgraphAdhCalculation}.

\begin{algorithm}
    \caption{Dep. Graph - Adhesions Calculation}
    \label{alg:depgraphAdhCalculation}
    \DontPrintSemicolon
    \small
    $\mathcal{G}^d \gets  \{ \}$ \tcp{Dep. Graph is a dictionary}
    $\mathcal{A} \gets  [ ]$\;
    \For{$B_i$ in $\mathcal{W}.\text{getBricks()}$} {
        \For{$B_j$ in $\mathcal{W}.\text{getBricks()}$} {
            \If{$B_i.\text{dependsOn}(B_j)$} {
                $start, width \gets \text{calcHorOverlap}(B_i, B_j)$\;
                $ad \gets  \text{Adhesion}(start, width)$\;
                $\mathcal{A}.\text{append}(ad)$\;
                $\mathcal{G}^d [B_i].\text{append}(ad)$\;
                $\mathcal{G}^d [ad].\text{append}(B_j)$\;
            }
        }
    }
    \Return $\mathcal{G}^d, \mathcal{A}$\;
\end{algorithm}

\subsection{Scheduling Constraints}
An overview of all the variables used throughout the problem formulation of the optimal scheduling problem along with their description is shown in Table \ref{tab:variables}.
\begin{table}
\centering
\caption{Formulation Variables of the Optimal Scheduling Problem}
\resizebox{0.95\linewidth}{!}{ 
\begin{tabular}{c c c}
    \hline
    \textbf{Variable}            & \textbf{Description}                          & \textbf{Type}      \\ 
    \hline
    $S^B_i$                      & Start time of task $B_i$                     & $\mathbb{R}^+$     \\ 
    $S^A_j$                      & Start time of task $A_j$                     & $\mathbb{R}^+$     \\ 
    $d^A_i$                      & Duration of task $A_j$                     & $\mathbb{R}^+$     \\ 
    $x_{i,k}$                    & Assignment of task $B_i$ to $R_k$            & $\{0,1\}$          \\ 
    $\alpha^\kappa_{i,j}$        & Order between $B_i, B_j$ assigned to $R_k$   & $\{0,1\}$          \\ 
    $\beta_{i,j}$                & Order between $A_i, A_j$                     & $\{0,1\}$          \\ 
    $\gamma^B_{i,j}$             & Order for conflicting tasks $B_i, B_j$       & $\{0,1\}$          \\ 
    $\gamma^{B,A}_{i,j}$         & Order for conflicting tasks $B_i, A_j$       & $\{0,1\}$          \\ 
    $y^A_{i,j}$                  & Precedence of $A_j$ to $A_i$                 & $\{0,1\}$          \\   
    $r^A_{i}$                    & Order of $A_i$ in time                       & $\mathbb{Z}^+$     \\ 
    $l^A_{i}$                    & Logistics time for $A_i$                     & $\mathbb{R}^+$     \\ 
    $C_{max}$                    & Mission makespan                             & $\mathbb{R}^+$     \\ 
    $s_{i}$                      & Slack variable for $A_i$                     & $\{0,1\}$    \\
    \hline
\end{tabular}
}
\label{tab:variables}
\end{table}
\subsubsection{Task Assignment}
Initially, the starting time of all tasks cannot be a negative number since the mission starts for $t=0$, thus the following constraints are added:
\begin{align}
    S^B_i & \geq 0, \quad \forall B_i \in \mathcal{B} 
    \label{eq:brickStart}\\
    S^A_i & \geq 0, \quad \forall A_i \in \mathcal{A}\label{eq:adhStart}
\end{align}
Given that the number of available pick-place robots is equal to $N$. the assignment of the adhesion tasks to them is handled via the matrix $\mathbf{X} \in \mathbb{B}^2$ where each binary element $x_{i,k} = 1$ if the task  $\mathcal{B}_i$ is assigned to the robot $R^B_k$ and $0$ otherwise. Each task $\mathcal{B}_i$ can be assigned to only one robot $R_k$ and this is handled by the classic constraint:
\begin{equation}
    \label{eq:brickAssign}
    \sum^N_{k=0} x_{i,k} = 1, \forall B_i \in \mathcal{B}
\end{equation}

\subsubsection{Tasks Ordering}
Furthermore, it is required that all tasks $B_i \in \mathcal{B}^k$, where $\mathcal{B}^k$ represents the set of bricks assigned to robot $R_k$, must be executed one at a time. Consequently, an ordering is enforced through the binary variable $\alpha^k_{i,j}$, that indicates the precedence relationship between a pair of tasks $(B_i, B_j) \in \mathcal{B}^k$ and is defined as follows:
\begin{equation}
    \label{eq:ordering1}
    \alpha^k_{i,j} =  
    \begin{cases} 
        1 & ,\text{if } S^B_i + d^B \leq S^B_j  \\
        0 & ,\text{otherwise }
    \end{cases}
\end{equation}
where $d^B$ denotes the duration of a brick task. This piecewise relation is transformed into linear constraints using the big-M method \cite{bigM_notation} 
\begin{equation}
    \label{eq:brickOrder}
    S^B_i + d^B \leq S^B_j + M(1 - \alpha^k_{i,j}),\: \forall (\mathcal{B}_i, \mathcal{B}_j )\in \mathcal{B}^k, R_k \in \mathcal{R}
\end{equation}
where $M \in \mathbb{R^+}$ is a number big enough such as $M > d^B_i$, $\forall d^B_i \in \mathcal{D}$.
In case two tasks $B_i, B_j$ are assigned to a robot $R^B_k$, the order between them can be defined by only one of the binary variables $\alpha^k_{i,j}$
since having both of them equal to $1$ would lead to ambiguity. Additionally, a robot can not execute two assigned tasks simultaneously, which leads to the constraint expressed by the following equation:
\begin{equation}
    \label{eq:ordering2}
    \alpha^k_{i,j} + \alpha^k_{j,i} = x_{i,k} \  x_{j,k},\: \forall  (\mathcal{B}_i, \mathcal{B}_j )\in \mathcal{B}^k, R_k \in \mathcal{R}
\end{equation}
Indicating that in case both tasks are assigned to the robot $R_k$ ($x_{i,k} \  x_{j,k}=1$), only one ordering variable will be equal to $1$ as well. 
Additionally, given the fact that there is only one adhesion spraying UAV, the adhesion tasks cannot be operated in parallel and a sequential order needs to be imposed. This ordering is handled using the binary variable $\beta_{i,j} \in \mathbf{B}$ which denotes that task $A_i$ precedes task $A_j$. This relation is expressed in a similar fashion with the above via the following constraints:
\vspace{-2mm}
\begin{equation}    
    \label{eq:adhesionOrdering}
    \forall (A_i, A_j) \in \mathcal{A}, A_i\neq A_j:
    \begin{cases}
        E^A_i  \leq S^A_j + M (1 - \beta_{i,j}) \\
        E^A_j  \leq S^A_i + M \beta_{i,j}
    \end{cases}
\end{equation}
where $E^A_i = S^A_i + d^A_i$ denotes the ending time of task $A_i$ and $d^A_i$ the duration of task $A_i$.

\subsubsection{Precedence}
As mentioned in Section \ref{sec:wallPlanGen}, the graph $\mathcal{G}^d$ captures all the dependencies between both tasks $B_i$ and $A_j$ so that the overall plan is completed successfully. These are transformed in precedence constraints where each edge $\mathcal{E} = (\mathcal{T}_i, \mathcal{T}_j)$ of the graph indicates that the source task $\mathcal{T}_i$ needs to be started after the completion of task $\mathcal{T}_j$, where $\mathcal{T}_i$,$\mathcal{T}_j$ are either brick or adhesion tasks. This constraint is incorporated for both two sets $\mathcal{P}^{A,B}$ and $\mathcal{P}^{B,A}$ as follows:
\begin{align}
    \label{eq:ABpreced}
    S^A_i + d^A_i& \leq S^B_j, \quad \forall (A_i, B_j) \in \mathcal{P}^{A,B}     \\
    \label{eq:BApreced}S^B_i + d^B & \leq S^A_j, \quad \forall (B_i, A_j) \in \mathcal{P}^{B,A}
\end{align}
where $\mathcal{P}^{A,B}$ and $\mathcal{P}^{B,A}$ are the sets of precedence constraints between adhesion and brick tasks and $d^A_i$ and $d^B$ are the duration of the adhesion and brick tasks respectively.

\subsubsection{Safety Consideration - Conflict Resolution}
Considering the working envelope of each robot, there is a potential risk of two robots operating in close proximity, which may lead to a collision during simultaneous execution. To prevent such occurrences, a minimum clearance $r_c \in \mathbb{R}$ between the UAVs must be maintained throughout the entire mission. This constraint is incorporated into the planning phase by computing the distances between all task locations and imposing concurrency constraints on tasks that violate the specified clearance requirement. Specifically, the different types of conflicts are identified and placed in sets, each one containing the conflicting tasks. There are two sets in total $\mathcal{C}^B_B, \mathcal{C}^A_B$ which are defined as follows:
\begin{align}
    \mathcal{C}^B_B = \{ (B_i, B_j) : \text{dist}(B_i, B_j)\leq r_c\} \\
    \mathcal{C}^A_B = \{ (A_i, B_j) : \text{dist}(A_i, B_j)\leq r_c\}
\end{align}
Having calculated the aforementioned sets, the pairs that have any kind of dependency between them are pruned so that fewer constraints are imposed.
%
For every pair, a concurrency constraint must be imposed so that the two tasks are not executed in parallel while the order between them is dynamically defined by the solver. The parallel operation can be easily detected and handled via the start time and duration of each task. On this, two binary variables $\gamma^B_{i,j}, \gamma^{B,A}_{i,j}, $ are added  indicating the order between the violating tasks $(B_i, B_j) \in {C}^B_B$ and $(B_i, A_j) \in {C}^B_A$. For each one of them, a pair of constraints is added similarly to Eq. \ref{eq:adhesionOrdering} as follows:
\begin{equation}
    \label{eq:conflBricks}
    \forall (B_i, B_j) \in \mathcal{C}^B :
    \begin{cases} 
        S^B_i + d^B \leq S^B_j + M (1 - \gamma^B_{i,j}) \\
        S^B_j + d^B \leq S^B_i + M \gamma^B_{i,j}
    \end{cases}
\end{equation}
\begin{equation}
    \label{eq:conflBrickAdh}
    \forall (B_i, A_j) \in \mathcal{C}^B_A :
    \begin{cases} 
        S^B_i + d^B  \leq S^A_j + M (1 - \gamma^{B,A}_{i,j})\\
        S^A_j + d^A_j \leq S^B_i + M \gamma^{B,A}_{i,j}
    \end{cases}
\end{equation}

\subsubsection{Adhesion Curing Time}
As discussed in Section \ref{sec:problStatement}, a key challenge in the construction mission is the curing time of the adhesive mortar between bricks. Once an adhesion task $A_i$ is completed, the mortar retains its adhesive properties for a limited time $d_{cur}$. Since the solver dynamically determines the start times for both adhesion and brick placement tasks, these constraints are modeled as dynamically coupled precedence deadline constraints. Once an adhesion task is completed, the corresponding brick placement must be completed within the curing time $d_{cur}$ to ensure proper bonding. This constraint is applied for all adhesion tasks $A_j$ as follows:
\begin{equation}
    \label{eq:curingConstr}
     S^B_i + d^B \leq S^A_j + d_{cur}, \; \forall B_i \in \mathcal{A}^A_j, \forall A_j\in \mathcal{A}
\end{equation}
where $\mathcal{A}^A_j = \text{anc}(A_j)$ is a set containing all the brick ancestors of the task $A_j$ in the dependency graph $\mathcal{G}^d$.

\subsubsection{Adhesion Robot Dynamic Logistics Time}
Since some adhesion tasks are positioned closely together while others may be arranged in an antisymmetric manner, it is essential to implement precise time scheduling and develop a dynamic time model for each task. This ensures a realistic representation of the travel time required between adhesion tasks but also facilitates the minimization of the travel time in between them (as explained later in Section \ref{sec:adhTravelMin}).
Aiming to model this, a ranking vector $\mathcal{R}$ is formulated where each element $r_i \in \mathbb{Z}^+$ denotes the ranking (order) of its corresponding adhesion task $A_i$ in time. It can be modeled as the column sum of the $i$-th column of matrix $\mathbf{B}$ and it can be conceived as a counter for the number of adhesion tasks succeeding the task $A_i$ and is defined as follows
\begin{equation}
    r_i = \sum^A_j \beta_{j,i}
\end{equation}

A method is developed to identify the next task in sequence after each task. Specifically, a task $A_j$ is considered the immediate successor of task $A_i$ if and only if $r_j - r_i = 1$.  

To enforce this condition, a binary variable $o_{i,j}$ is introduced to indicate whether $A_j$ is the immediate successor of $A_i$ and is defined as follows:  
\begin{equation}
    o_{i,j} = 
    \begin{cases}
        1, & \text{if} \: r_j - r_i = 1 \\
        0, & \text{otherwise}
    \end{cases}
\end{equation}

To incorporate this condition into an optimization framework, the big-M method is used:  

\begin{equation}
 \label{eq:adhesionRanking}
 \forall (A_i, A_j), A_i \neq A_j:
    \begin{cases}
        r_j - r_i \leq 1 + M(1 - o_{i,j}) \\
        r_j - r_i \geq 1 - M(1 - o_{i,j})
    \end{cases}
\end{equation}

However, these constraints alone do not fully define $o_{i,j}$. To ensure that each task is followed by exactly one other task, an additional constraint is imposed. This constraint enforces that the row sum of each task, representing the number of immediate successors, must be equal to 1:  
\begin{equation} \label{eq:adhesionTaskOrderingUnmod}
    \sum_{j} o_{i,j} = 1, \quad \forall A_i \in \mathcal{A}
\end{equation}
This guarantees that each task has a unique immediate successor, maintaining a properly ordered sequence.  However, the final task $A_f$ in time is not followed by any other one indicating that its row sum is equal to zero. Thus, aiming to handle this properly, without compromising the flexibility of the algorithm, a slack binary vector $\mathbf{s} = \{s_i, \forall A_i \in \mathcal{A} \}$ is added and the constraint of Eq. \ref{eq:adhesionTaskOrderingUnmod} is transformed as follows:
\begin{equation} \label{eq:adhesionTaskOrderingMod}
    \sum_{j} o_{i,j} = 1 - s_i, \quad \forall A_i \in \mathcal{A}
\end{equation}
where the flexibility is provided to dynamically select which task $A_f$ will be the last one by setting the corresponding slack variable $s_f =1$. However, only one task must be the final one so an additional constraint is added implying that only one of the slack variables must be equal to $1$ as follows:
\begin{equation}
    \label{eq:slackVariableRow}
    \sum^A_i s_i =1, \; \forall A_i \in \mathcal{A}
\end{equation}

Thus, now that the variable $o_{i,j}$ is sufficiently defined, the dynamic logistics time for the adhesion tasks can be defined as follows:
\begin{equation}
    t^A_i = \sum^A_{i=0} o_{i,j} \: l^A_{i,j}/v_{log}
\end{equation}

where $l^A_{i,j} = dist(A_i, A_j)$ denotes the distance between the adhesion tasks $A_i$ and $A_j$ and $v_{log}$ is constant corresponding to the speed of the UAV while operating logistics movement from one adhesion chunk to another. Finally, the total adhesion task duration $d^A_i$ is defined as $d^A_i = d_s + t^A_i$ where the $d_s \in \mathbb{R}$ is constant and denotes the duration of spraying mortar for a single adhesion task.

\subsection{Scheduling Optimization Objectives}

\subsubsection{Makespan Minimization}
The main objective of the scheduling is to minimize the makespan $C_{\text{max}}$ of the overall mission, which is defined as the maximum completion time of all tasks to be handled. Given the nature of the mission, the last task that is going to be completed will be a brick type, thus, the makespan $C_{\text{max}}$ is defined as follows:
\begin{equation}
    \label{eq:makespanCon}
    S^B_i + d^B \leq C_{\text{max}}, \quad \forall B_i \in \mathcal{B} 
\end{equation}
The makespan $C_{\text{max}}$ is minimized by incorporating it in the overall objective function.

\subsubsection{Pick/Place Robots Logistics Minimization}
The goal is to minimize the distance traveled between the robot's $R^B_k$ pickup point $P_k$ and the placement position of each brick task $B_i$.
Thus, a distance matrix $\mathbf{L}^B$ is computed, where each element $l^B_{i,k} = \text{dist}(B_i, P_k)$ represents the distance between brick task $B_i$ and its corresponding pickup point $P_k$. The efficiency of the robots’ logistics is then quantified using the following score:
\begin{equation}
    J^B_{log} = \sum^B_{i} \sum^N_{k} x_{i,k} l^B_{i,k}
\end{equation}

\subsubsection{Curing Time minimization}
To enhance the structural integrity and bonding strength between bricks, the time delay between placing an adhesion material $A_i$ and its corresponding brick $B_i$, which is placed on top, apart from being upper bounded (Eq. \ref{eq:curingConstr}) is also minimized. This ensures that the adhesion material remains fresh and results in more efficient bonding, as its adhesiveness decays over time. The objective is formulated as follows:
\begin{equation}
    J_{cur} = \sum^A_i\sum_{b \in \mathcal{A}^A_i} \big(  S^B_j - (S^A_i + d^A_i) \big) 
\end{equation}
where  $\mathcal{A}^A_i$ is a set containing all the brick ancestors of the task $A_i$ in the dependency graph $\mathcal{G}^d$.

\subsubsection{Adhesion Robot Logistics Minimization}\label{sec:adhTravelMin}

When executing adhesion tasks sequentially, minimizing both time and distance is crucial to reduce UAV movements and ensure smooth mortar extrusion. A UAV that has completed adhesion tasks should prioritize nearby ones, especially if they can be executed continuously without pausing the spray action. To capture this, the adhesion logistics score $J^A_{log}$ is computed, representing the total travel distance for the adhesion UAV. The pairwise task distance is given by $l^A_{i,j} = \text{dist}(A_{i}, A_{j})$, where $\text{dist}(\cdot,\cdot)$ denotes the Euclidean distance.

However, the order of the tasks is determined dynamically during optimization. To capture this, the binary ordering variable $o_{i,j}$ is used, and the adhesion logistics score is defined as follows:
\begin{equation}
    J^A_{log} = \sum^A_{i=0} \sum^A_{j=0} l^A_{i,j} \: o_{i,j}
\end{equation}

\subsection{Problem Overview}
Finally, all the aforementioned objective scores are scaled and added in a common cost function $J$ as follows:
\begin{equation}
    J = W_{span} \: C_{max} + W^B_{log} \: J^B_{log} + W_{cur} \: J_{cur} +  W^A_{log} \: J^A_{log}
\end{equation}
where $ W_{span}, W^B_{log}, W_{cur}$ and $W^A_{log}$ are tunable positive gains. An overview of the whole problem formulation is given as follows:
%
\begin{equation}
\begin{aligned}
\argmin_{} \quad & J
\\
\textrm{subject to:}\\ 
    \forall \mathcal{A}_i \in \mathcal{A} :& \quad  \text{Eq. (\ref{eq:adhStart}), (\ref{eq:adhesionTaskOrderingMod}), (\ref{eq:slackVariableRow})} \\
    \forall \mathcal{B}_i \in \mathcal{B} :& \quad  \text{Eq. (\ref{eq:brickStart}), (\ref{eq:brickAssign}), (\ref{eq:makespanCon})} \\
    \forall (\mathcal{B}_i, \mathcal{B}_j )\in \mathcal{B}^k :& \quad  \text{Eq. (\ref{eq:brickOrder}) }\\
    \forall (A_i, B_j) \in \mathcal{P}^{A,B}    : &\quad \text{Eq. (\ref{eq:ABpreced})} \\
    \forall (B_i, A_j) \in \mathcal{P}^{B,A}    : &\quad \text{Eq. (\ref{eq:BApreced})} \\
    \forall (B_i, B_j) \in \mathcal{C}^B_B        :& \quad \text{Eq. (\ref{eq:conflBricks})}  \\
    \forall (B_i, A_j) \in \mathcal{C}^B_A      :& \quad \text{Eq. (\ref{eq:conflBrickAdh})} \\
    \forall B_i \in \mathcal{A}^A_j, \forall A_j\in \mathcal{A}:& \quad \text{Eq. (\ref{eq:curingConstr})} \\
     \forall (A_i, A_j) \in  \mathcal{A}:& \quad \text{Eq. (\ref{eq:adhesionOrdering}), (\ref{eq:adhesionRanking})} 
\end{aligned}
\end{equation}

The problem falls in the category of Mixed Integer Programming (MIP) and its formulation along with its solution is carried out using the Gurobi Optimization solver \cite{gurobi}.

\subsection{Mission Execution}
\subsubsection{Mission Control}
The mission planner receives the schedule once it has been optimized based on the mission requirements and the UAVs' configuration. By defining the timing for all tasks $B_i, A_j$ and their corresponding robots $R^B_k, R^A$, the mission control transforms it into a time-based event log.
%
%
Tasks are initiated in accordance with the timetable as time progresses. The mission control allocates the associated job to the assigned agent when the time for a new task starts, according to the plan.

\subsubsection{Collision Avoidance During Logistics}
While traveling between the home positions and the construction
workspace, UAVs fly at a cruising altitude $h_{cr}$ and collision avoidance during logistics movements is handled in a decentralized way by UAVs sharing their planned trajectories, similar to the approach in \cite{bjornCollisionAvoidance}. Each UAV treats the trajectories of others as obstacles in space-time and re-plans its own trajectory in case a potential collision is detected.

\subsubsection{Brick Pick-Place UAV}
Each brick pick-place UAV,$R^B_k$, is equipped with a gripper mounted on its underside. 
%
%
When a robot $R^B_k$ is assigned a brick task, $B_i$, it first takes off if it is currently on the ground. The UAV then navigates to its designated pickup point, $P_k$, to retrieve the brick. After pick up, it ascends to the cruising altitude, $h_{cr}$, and flies to the placement position. Upon reaching this location, the UAV hovers briefly to stabilize itself before beginning its descent. Once it is close enough to the placement position and stable, it releases the brick. Following this, the UAV ascends back to the cruising altitude and either returns to the pickup point if assigned another brick task or heads to its home position to land.

\subsubsection{Adhesion UAV}
Similarly, the adhesion UAV, $R^A$, is equipped with an adhesion material extruder mounted on its underside, along with a canister containing the material. When assigned an adhesion task, $A_j$, the UAV hovers at the cruising altitude, $h_{cr}$, and moves to the task's starting point. After stabilizing, it begins spraying mortar smoothly between the bricks along the entire span of the task. Once the task is complete, the UAV either proceeds to the next task or lands at a predefined position near the wall.

\section{Results}
\begin{figure*}
    \centering
    \includegraphics[width=0.95\linewidth]{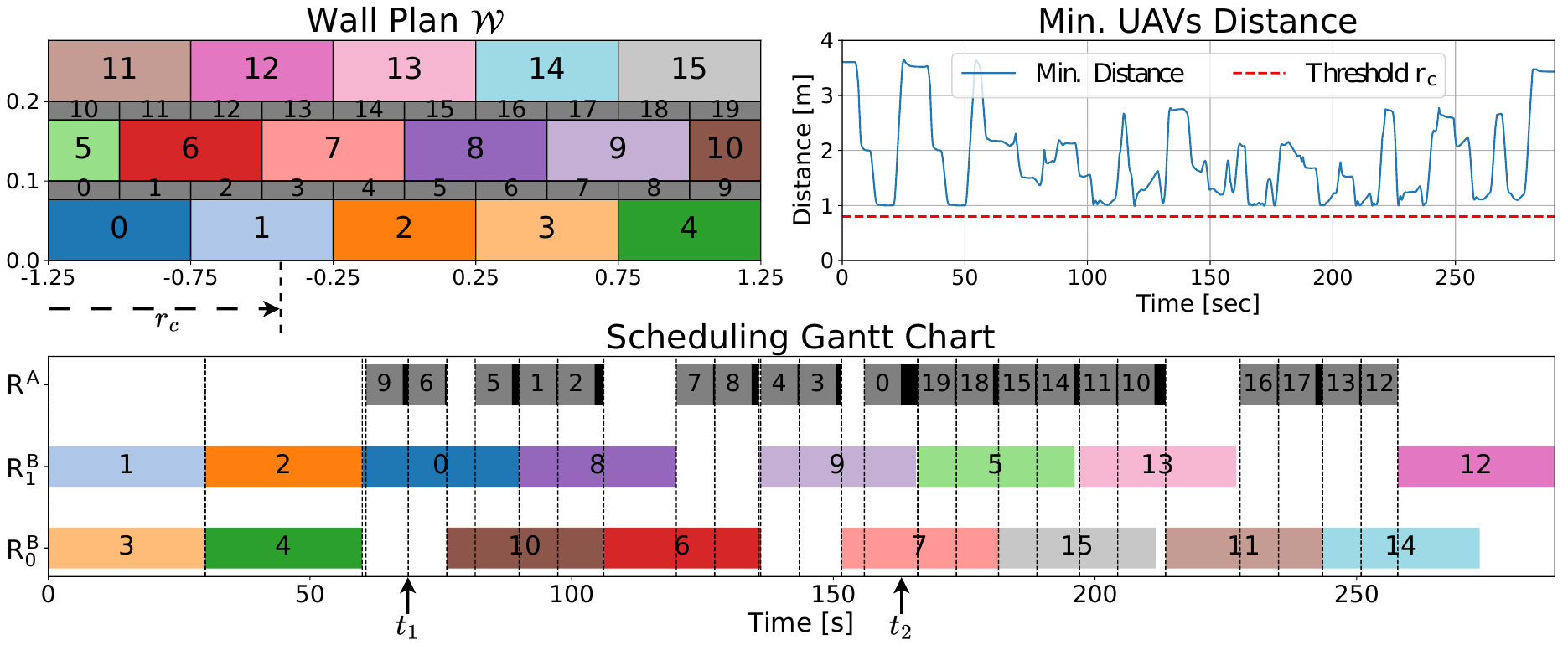}
    \vspace{-4mm}
    \caption{The computed plan $\mathcal{W}$ for a $2.5 \times 0.3$ m wall illustrates each brick task $B_i$ in color and mortar tasks in gray (top left). The minimum distance between all UAVs is shown in blue, along with the minimum clearance $r_c = 0.8$ m in red (top right). The optimal schedule is then visualized in a Gantt Chart format, indicating the start time of each task and the robot it's assigned to. The dynamic logistics time of adhesion is colored with black (bottom).}
    \label{fig:caseStudyWallPlanGantt}
    \vspace{-5mm}
\end{figure*}

To evaluate the proposed framework in a robotic construction mission, a Gazebo simulation environment is used, where two brick pick-and-place UAVs, $R^B_1$ and $R^B_2$, along with an adhesion UAV, $R^A$, are simulated and modeled based on \cite{carma}. Since the primary focus is on optimal planning and scheduling, the physics of brick contact and adhesion material deposition are considered out of scope of this paper. Instead, it is assumed that each brick $B_i$ is picked up and placed by the robots $R^B_k$ as soon as their gripper reaches sufficiently close to their position. Additionally, the extruder tip is tracked in 3D, and the mortar spraying process is represented by small spherical markers placed at a vertical offset from the extruder whenever a spraying action is commanded. A case study of a rectangular wall of length and width equal to $w_m = 2.5$ m. and $h_m=0.3$ m. correspondingly is evaluated with brick dimensions of $w_{fb} = 0.5$ m. and $h_{fb} = 0.1$ m. Given the fact that the wall is going to be centered at the origin of the workspace, the two brick pickup points corresponding to the two UAVs $R^B_0$ and $R^B_1$ are placed as follows $P_0 = [1.5, -1.0]$ m. and $P_1 = [-1.5, -1.0]$ m.  

The automated wall plan $\mathcal{W}$ generated from the bricks plan generator module contains $3$ layers of bricks and is composed of $16$ brick tasks $B_i$ and $20$ adhesion tasks $A_j$ and is shown at the top part of Fig. \ref{fig:caseStudyWallPlanGantt}. The bricks are color-coded in order to be easily distinguishable while the adhesion tasks are visualized with gray to resemble the mortar. It is noticed that there are more than one adhesion tasks $A_j$ per brick $B_i$ and this is done intentionally to provide the solver with greater flexibility when it comes to parallel execution of neighboring tasks.

The dependency graph $\mathcal{G}^d$ is calculated along with the conflicting sets $\mathcal{C}^B_B$ and $\mathcal{C}^B_A$ where a minimum clearance of $r_c = 0.8$ m. is considered.
The durations of the brick task and adhesion spraying along with mortar curing time are set to $d_p = 30$, $d_s=7$ and $d_c = 60$ sec, while the logistics speed is set to $v_{log} = 0.6$ m/s. The gains of the objectives are selected as follows
$W_{sp} = 2, W^B_{log} = 4, W^A_{log} = 5, W_{cur} =0.2$.
%
%

\begin{figure*}
    \centering
    \includegraphics[width=\linewidth]{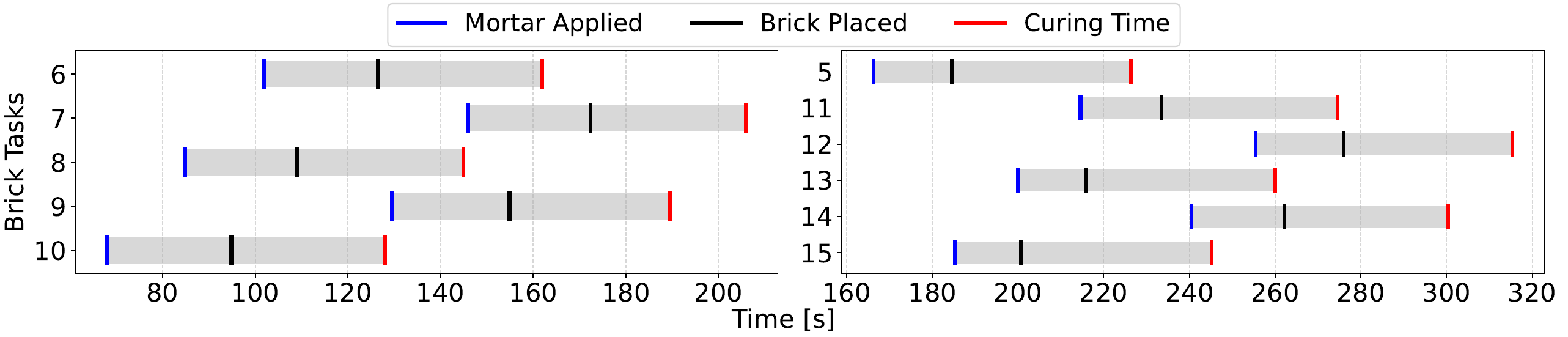}
    \vspace{-8mm}
    \caption{Mortar curing time windows for brick tasks $B_i$, where the time the mortar is applied (blue), the curing window (gray), the curing time (red), and the actual time the brick is placed (black) are shown.  }
    \label{fig:curingWindows}
    \vspace{-5mm}
\end{figure*}

%
The computed optimal schedule $O_{sc}$ is illustrated in the lower section of Fig. \ref{fig:caseStudyWallPlanGantt} where both adhesion and brick tasks are illustrated in a color-coded format and indicates both their starting and ending time resulting in overall makespan of the whole mission equal to $C_{max} = 290$ sec. 
Initially, the first four bricks $B_1 -B_4$ are assigned to available robots $R^B_0, R^B_1$ and prioritized before any adhesion task begins, ensuring that the static structural and geometric constraints captured in the dependency graph are respected. This initial selection provides the solver with greater flexibility in task allocation later. Right after the completion of $B_1, B_3$, the adhesion task $A_9$ is assigned.
Additionally, the spatio-temporal minimum clearance constraints are enforced, this is highlighted for $t=t_1= 68$ sec.,  where although the completion of task $A_9$ has been completed—making the placement of brick $B_{10}$ theoretically possible from a structural perspective—
the task is intentionally delayed. This is because task $B_{10}$ is in conflict with $A_6$ (since the distance between the two tasks is less than the required clearance $r_c$), hence they cannot be executed at the same time. As a result, $B_{10}$ is slightly delayed to maintain sufficient separation between $R^A$ and $R^B_0$. This is made possible by providing flexibility to the adhesion tasks $A_6$ and $A_5$ and not merging them into a common task, despite both referring to the same brick.
The scheduled plan not only reduces the overall mission makespan but also optimizes the timing between adhesion application and the subsequent brick placement. This is particularly evident at $t = 150$ sec, where task $A_0$ could have been executed immediately after $A_3$. However, it is slightly delayed to synchronize its completion with that of $B_9$. This ensures that immediately afterward, $R^B_1$ can place brick $B_5$, while $R^B_0$ remains engaged with another task in the meantime.
The dynamic adhesion logistics time is also evident for $t = t_2 = 162$ sec. where the transition from $A_0$ to $A_{19}$ takes considerably greater duration compared to the other ones.
Throughout the simulated mission, the minimum distance between all three UAVs is continuously monitored and plotted over time, as shown in the top right section of Fig. \ref{fig:caseStudyWallPlanGantt}. The results confirm that the minimum clearance requirement of $r_c = 0.8$ m is consistently maintained.
This adherence to safety constraints is ensured by the scheduling approach, which incorporates conflict avoidance measures. Additionally, any potential conflicts arising during logistics are resolved in real time through the trajectory-sharing scheme.
Finally, for each brick, the completion time of its associated mortar task and the time at which the brick is placed are tracked and visualized in Fig. \ref{fig:curingWindows}. The required placement window is represented in gray, indicating the time-frame within which the brick must be placed. The results show that all brick placements occur within this window, demonstrating the effectiveness of the scheduling approach. It is also observed that, in the majority of cases, the brick is placed within the first half of the window. This is due to the minimization objective of curing time, ensuring that the mortar retains most of its adhesive properties and enhances overall construction quality. A video of the simulated mission is available at \href{https://youtu.be/kGvFGDCUkDQ}{https://youtu.be/kGvFGDCUkDQ}.

\section{Conclusion}
In this paper, a novel scheduling framework for autonomous robotic aerial masonry construction is presented, aiming to enhance the efficiency and coordination of multiple heterogeneous UAVs. The framework specifically addresses the challenge of mortar placement, considering the limited curing time window during which the mortar can effectively bond two bricks after being applied. Additionally, both static constraints corresponding to structural requirements, and spatio-temporal safety constraints are generated with the proposed automated pipeline which ensures safe and efficient autonomous heterogeneous construction.
The effectiveness of the proposed framework has been validated through Gazebo simulations, demonstrating its potential for practical applications in aerial masonry construction. Future work will focus on refining the framework for large-scale scenarios while exploring additional complexities associated with its real-life execution in the field. This research contributes to advancing the field of robotic construction, paving the way for more efficient and adaptable building methods utilizing UAV technology.

\vspace{-2mm}
\bibliographystyle{IEEEtran}
\bibliography{sample}

\end{document}